\documentclass[a4paper,10pt]{article}

\usepackage[margin=25mm]{geometry}
\usepackage{times}
\usepackage{graphicx}
\usepackage{caption}
\usepackage{amsmath}
\usepackage{dsfont}
\usepackage{xcolor}
\usepackage{fancyhdr}
\newcommand{\bx}{\boldsymbol{x}}
\newcommand{\balpha}{\boldsymbol{\alpha}}
\newcommand{\given}{\,|\,}
\newcommand{\blambda}{\boldsymbol{\lambda}}
\newcommand{\revision}{} 

\usepackage{titlesec}
\titleformat{\section}{\bfseries\large\uppercase}{\thesection.}{1em}{}
\titleformat{\subsection}{\bfseries\normalsize}{\thesubsection}{1em}{}
\titleformat{\subsubsection}{\itshape\normalsize}{\thesubsubsection}{1em}{}

\usepackage[numbers]{natbib}

\begin{document}

\title{\bfseries Bayesian Hierarchical Models and the Maximum Entropy Principle}
\author{Brendon J. Brewer$^{1}$ \\
{\small $^{1}$Department of Statistics, The University of Auckland, bj.brewer@auckland.ac.nz}}
\date{}
\maketitle
\thispagestyle{fancy}

\begin{abstract}
Bayesian hierarchical models are frequently used in practical data
analysis contexts.
One interpretation of these models is that they provide an indirect way
of assigning a prior for unknown parameters, through the introduction
of hyperparameters. The resulting marginal prior for the parameters (integrating
over the hyperparameters)
is usually dependent, so that learning
one parameter provides some information about the others.
In this contribution, I will demonstrate that,
{\revision when the prior given the hyperparameters is a canonical
distribution (a maximum entropy distribution with moment constraints),
the dependent marginal prior also has a maximum entropy property,
with a different constraint. 
This constraint is on the {\em marginal distribution} of some function of 
the unknown quantities}.
The results
shed light on what information is actually being assumed
when we assign a hierarchical model.
\end{abstract}

\section{Introduction}
Consider a collection of unknown quantities
\begin{align}
\bx = \{x_1, x_2, ..., x_n\}
\end{align}
for which we want to assign a joint probability distribution.
These quantities could be data values that will eventually become known,
or they might be parameters --- the important thing is that they are
currently unknown.
A simple choice of prior that describes a large amount of ignorance
is the uniform prior, 
\begin{align}
\pi(\bx) &\propto 1,
\end{align}
defined over some suitably wide domain of possible values of $\bx$.
However, when considering some functions $T_i = f_i(\bx)$ that are of particular
interest, the prior for the $T$ values implied by $\pi$ is often unsuitable.
If the expected values of the $\{T_i\}$ are specified (somehow),
the prior may be updated to take
into account this constraint using the principle of maximum entropy
\citep{jaynes, caticha, caticha_lectures}, giving the well-known result
\begin{align}
p(\bx) &= \frac{\pi(\bx)\exp\left(\sum_i\lambda_i f_i(\bx)\right)}{Z(\boldsymbol{\lambda})}.\label{eqn:canonical}
\end{align}
The values of the $\lambda$s need to be selected so as to enforce the desired
values of $\left<T_i\right>$. The result is the `canonical' family of
distributions from statistical mechanics \citep{jaynes1957}.

However, it would be unusual for expected values (which are properties
of the probability distribution, not of $\bx$) to really be `known'.
If they are unknown, it is tempting to use the canonical distribution
of Equation~\ref{eqn:canonical} as if it were the distribution {\em conditional}
on those expected values only. A marginal distribution for the
expected values would complete the model specification, and the resulting
marginal prior for $\bx$ could be found by integrating out the hyperparameter.
{\revision This marginal distribution would be a mixture of canonical distributions.}
However, the maximum entropy interpretation is {\revision apparently} lost,
because a mixture of canonical distributions is not itself a canonical distribution.
{\revision The purpose of this paper is to clarify that a maximum entropy
interpretation of the marginal prior for $\bx$ does in fact exist, and to identify
what the effective constraint is.}

\section{Bayesian Hierarchical Models}
In a Bayesian inference context, a prior distribution for a collection of
quantities $\bx$ would often be
assigned hierarchically, especially when there is prior information
connecting the $\{x_i\}$ together. With the introduction of hyperparameters
$\balpha$, the prior is defined in two stages, with a prior $p(\balpha)$
for the hyperparameters and then a conditional prior
$p(\bx \given \balpha)$ for the parameters given the hyperparameters.
The conditional prior is often the product of $n$ `iid' distributions,
one for each of the $x$s:
\begin{align}
p(\bx \given \balpha) &= \prod_{i=1}^n p(x_i \given \balpha).
\end{align}

The implied marginal prior for $\bx$ can be found by integrating
over the hyperparameters:
\begin{align}
p(\bx) &= \int p(\balpha)p(\bx \given \balpha) \, d\balpha \\
       &= \int p(\balpha)\prod_{i=1}^n p(x_i \given \balpha) \, d\balpha.\label{eqn:marginal}
\end{align}
Such a choice is often justified by exchangeability ---
when the $\{x_i\}$ values need to be treated interchangeably, the
hierarchical construction guarantees it. In the limit of large $n$,
the argument also works the other way around, in that the joint 
distribution for any exchangeable sequence of quantities $\{x_i\}$
can be expressed as in Equation~\ref{eqn:marginal}.
However, since the principle of maximum entropy is rarely invoked in
justifying such a choice, it is unclear whether there is any connection
between the maximum entropy principle and the choice of the hierarchical
model.

\section{Maximum Entropy Constraints}
The most common type of constraint used in maximum entropy is
constraints on expected values. However, in principle, any constraint
on the probability distribution (so-called {\em testable information})
can be used \citep{stand_on_entropy}.
Here, we will identify the implicit constraint that leads to
{\revision results like}
Equation~\ref{eqn:marginal} {\revision for} the updated maximum entropy distribution
for $\bx$.

Consider a prior $\pi(\bx)$ and suppose that some
function of $\bx$, written as $T = f(\bx)$, is of particular interest.
In {\revision many} cases, the prior over $\bx$ would imply an inappropriate
prior for $T$. For example, suppose that $\pi$ is flat over the
range $[0, 100]$ for each $x_i$, and that $T = \frac{1}{n}\sum_i x_i$.
By the central limit theorem, the implied prior for $T$ would be
approximately
\begin{align}
T &\sim \textnormal{Normal}\left(50, \left(\frac{100}{\sqrt{12n}}\right)^2\right),
\end{align}
which could be narrow for large $n$.
{\revision Of course, in many cases it is appropriate and desirable for
certain distributions to be narrow, but here the assumption is that this is
an unintended consequence of the naive uniform prior}.
We revisit this example in Section~\ref{sec:exponential}.

To fix this, we could apply a constraint to the {\em implied marginal distribution of $T$},
so that it does what we want.
To see the resulting MaxEnt distribution,
{\revision we will now examine a simple case, from which we can infer the
general result}.
Consider a case where $T=f(\bx)$
could only take values 1, 2, and 3.
{\revision According to any distribution $p(\bx)$,}
The probability that $T=1$ would be given by
\begin{align}
P(T=1) &= \int p(\bx) \mathds{1}\left(f(\bx) = 1\right) \, d\bx,
\end{align}
{\revision where $\mathds{1}()$ is an indicator function that returns
1 if the argument is true, and 0 if it is false.}
Similarly, the probabilities $P(T=2)$ and $P(T=3)$ would be
\begin{align}
P(T=2) &= \int p(\bx) \mathds{1}\left(f(\bx) = 2\right) \, d\bx \\
P(T=3) &= \int p(\bx) \mathds{1}\left(f(\bx) = 3\right) \, d\bx.
\end{align}
These three values would form the implied probability distribution for
$T$ {\revision according to $p(\bx)$}.
However, the expressions for the three probabilities are also
interpretable as expected values (of indicator functions).
Therefore, controlling the implied marginal distribution of $T$
is the same as controlling three expected values, and the MaxEnt solution
is the canonical distribution:
\begin{align}
p(\bx) &\propto \pi(\bx)\exp\left(\lambda_1
\mathds{1}\left(f(\bx) = 1\right) +
\lambda_2
\mathds{1}\left(f(\bx) = 2\right) +
\lambda_3
\mathds{1}\left(f(\bx) = 3\right)
\right).
\end{align}
As usual, the $\lambda$ hyperparameters would have to be chosen so that
the desired prior for $T$ is obtained.

Note that the expression inside the exponential is simply a way
of expressing a transformed version of $T$, called the indicator
function expansion. If $T=1$ the function
returns $\lambda_1$, if $T=2$ it returns $\lambda_2$, and if
$T=3$ it returns $\lambda_3$. We can therefore write the result more simply
as
\begin{align}
p(\bx) &\propto \pi(\bx)\exp\left(g(f(\bx))
\right).
\end{align}
where $g()$ is a function that needs to be tweaked until the desired
marginal distribution of $T$ is achieved.
The exponential can be absorbed into the function $g()$, giving the
simpler result
\begin{align}
p(\bx) &\propto \pi(\bx)g\left(f(\bx)\right),\label{eqn:result}
\end{align}
where the function $g()$ can no longer be negative.
While motivated by the simple situation with only three possible values
for $T$, Equation~\ref{eqn:result} is in fact the general solution for a single
derived quantity $T$.

If there are several such derived quantities $T_1, ..., T_m$
given by functions $f_1(\bx), ..., f_m(\bx)$, the {\revision general expression
for the updated probability distribution $p(\bx)$}
becomes
\begin{align}
p(\bx) &\propto \pi(\bx)g\left(f_1(\bx), f_2(\bx), ..., f_m(\bx)\right).
\label{eqn:result2}
\end{align}

{\revision
We are now in a position to show that mixtures of canonical distributions
are of the form of Equation~\ref{eqn:result2}, and are therefore maximum
entropy distributions with a constraint on the marginal distribution of a
derived quantity. Let $\bx$ be the unknown quantities, and assume that
the conditional distributions are canonical distributions:
\begin{align}
p(\bx \given \blambda) &= \frac{\pi(\bx)\exp\left(\sum_i\lambda_i f_i(\bx)\right)}{Z(\blambda)}.
\end{align}
This equation is equivalent to Equation~\ref{eqn:canonical} except the dependence
on $\blambda$ is now explicit on the left hand side.

The marginal distribution for $\bx$ can be found by assigning a prior
for the Lagrange multipliers, and then integrating them out:
\begin{align}
p(\bx) &= \int p(\blambda)p(\bx \given \blambda) \, d\blambda \\
       &= \int p(\blambda)\frac{\pi(\bx)\exp\left(\sum_i\lambda_i f_i(\bx)\right)}{Z(\blambda)} \, d\blambda.
\end{align}
We do not need to perform this integral to see that overall, the expression
only depends on $\bx$ through the functions $\{f_i(\bx)\}$, which act
as `sufficient statistics'.
The marginal distribution $p(\bx)$ is therefore of the form
of Equation~\ref{eqn:result2}, and is a maximum entropy distribution,
where the prior was $\pi(\bx)$ and the
constraint for the updating procedure is a specified distribution for the
derived quantities $\{f_i(\bx)\}$. The maximum entropy update applies only
on the space of $\bx$ (not the joint space of $\bx$ and $\lambda$),
and the Lagrange multiplier hyperparameters
can be interpreted as a practical device to make
the whole procedure tractable.

We will now examine two elementary examples.
}

\section{Exponential Example}\label{sec:exponential}
As a simple example, consider a flat prior $\pi$ defined between
0 and 100 for some positive quantities $\bx = \{x_1, ..., x_n\}$.
As discussed previously, the central limit theorem gives (to a good
approximation) the implied prior
for the arithmetic mean $T = \frac{1}{n}\sum_{i=1}^{n} x_i$.
The result
is a normal distribution, which might be quite narrow,
{\revision and this might be an unintended consequence of the uniform
distribution rather than a justified conclusion}.
For example, if $n=100$ the implied prior for $T$ is
\begin{align}
T &\sim \textnormal{Normal}\left(50, 2.89^2\right).
\end{align}

If we were to specify a constraint on the expected
value of $T$,
\begin{align}
\left<\frac{1}{n}\sum_{i=1}^{n} x_i\right> &= \mu,
\end{align}
then assuming that $\mu \ll 100$,
the updated distribution via MaxEnt, for this moment constraint,
would be a product of independent exponentials:
\begin{align}
p(\bx) &= \prod_{i=1}^{n} \frac{1}{\mu} \exp\left(-\frac{x_i}{\mu}\right) \\
       &\propto \mu^{-n} \exp\left(-\frac{nT}{\mu}\right).
\end{align}
If we let the
constraint value $\mu$ be unknown, this is the conditional distribution
$p(\bx \given \mu)$ only. However, we can find the marginal
distribution by integrating out $\mu$:
\begin{align}
p(\bx) &= \int_0^\infty p(\mu) \mu^{-n} \exp\left(-\frac{nT}{\mu}\right) \, d\mu.
\end{align}
This is of the form of Equation~\ref{eqn:result}, since the integral depends
on $\bx$ only through $T$. Therefore, this is a maximum entropy distribution
for $\bx$ with a constraint on the marginal distribution of $T$.

This argument implies that applying MaxEnt with a moment constraint to
obtain a conditional distribution, then applying a prior over the hyperparameter
in the conditional distribution, is equivalent to applying MaxEnt with
a constraint on the {\em marginal distribution} of $T$.

The difficulty with this approach is
that it is hard to find the function $g()$
that would lead to a particular marginal distribution of $T$. By using
the hierarchical route, we are indirectly controlling the prior for $T$
by choosing the prior for the hyperparameter $\mu$. Hopefully, for some
choice of prior $p(\mu)$, we can obtain an acceptable prior over $T$,
and in doing so, we will end up with a distribution over $\bx$ that
is a maximum entropy distribution.

\begin{figure}
\centering
\includegraphics[width=0.6\textwidth]{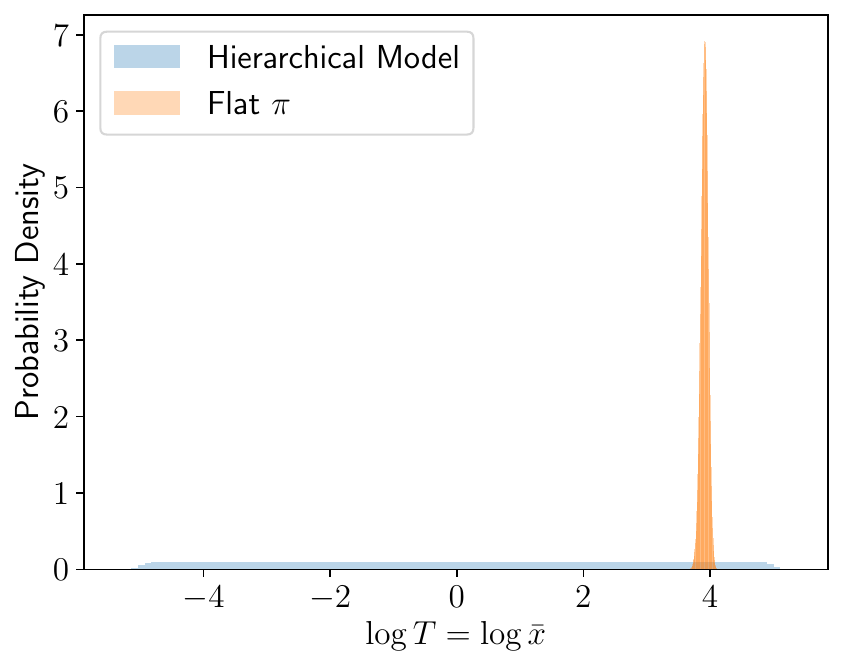}
\caption{The orange distribution is the implied prior for the log of the arithmetic mean of 100 positive quantities, using a Uniform$(0, 100)$ distribution.
The blue distribution is the implied prior using a hierarchical
model with $\log\mu \sim \textnormal{Uniform}(-5, 5)$, expressing
more appropriate prior uncertainty about the arithmetic mean.
\label{fig:xbar_prior}}
\end{figure}

In this specific example,
a common choice would be to use a log-uniform prior for the hyperparameter
$\mu$. For example, if we use $\log\mu \sim \textnormal{Uniform}(-5, 5)$,
Figure~\ref{fig:xbar_prior} shows the resulting implied prior distribution
for $T$, which is close enough to log-uniform for practical purposes.

\section{Gaussian Example}
Consider $n$ values $x_1, x_2, ..., x_n$ which may be positive or negative. Let the prior
$\pi(\bx)$ be flat over a very wide range. Suppose that we are interested in the following two
quantities derived from $\bx$:
\begin{align}
T_1 &= f_1(\bx) = \sum_{i=1}^n x_i \\
T_2 &= f_2(\bx) = \sum_{i=1}^n x_i^2.
\end{align}

If we fixed the expected values of $T_1$ and $T_2$, the maximum entropy
distribution is:
\begin{align}
p(\bx) &\propto \exp\left(\lambda_1 \sum_{i=1}^n x_i + \lambda_2 \sum_{i=1}^n x_i^2\right) \\
    &= \exp\left(\sum_{i=1}^n \left[\lambda_1 x_i + \lambda_2 x_i^2\right]\right),
\end{align}
which is a product of $n$ `iid' normal distributions and may also be written
as
\begin{align}
p(\bx \given \mu, \sigma)
    &\propto \prod_{i=1}^n
            \exp\left(-\frac{1}{2\sigma^2}\left(x_i - \mu\right)^2\right),
\end{align}
where the more traditional hyperparameters $\mu$ and $\sigma$ replace the
Lagrange multipliers $\lambda_1$ and $\lambda_2$.

\begin{figure}
\centering
\includegraphics[width=0.6\textwidth]{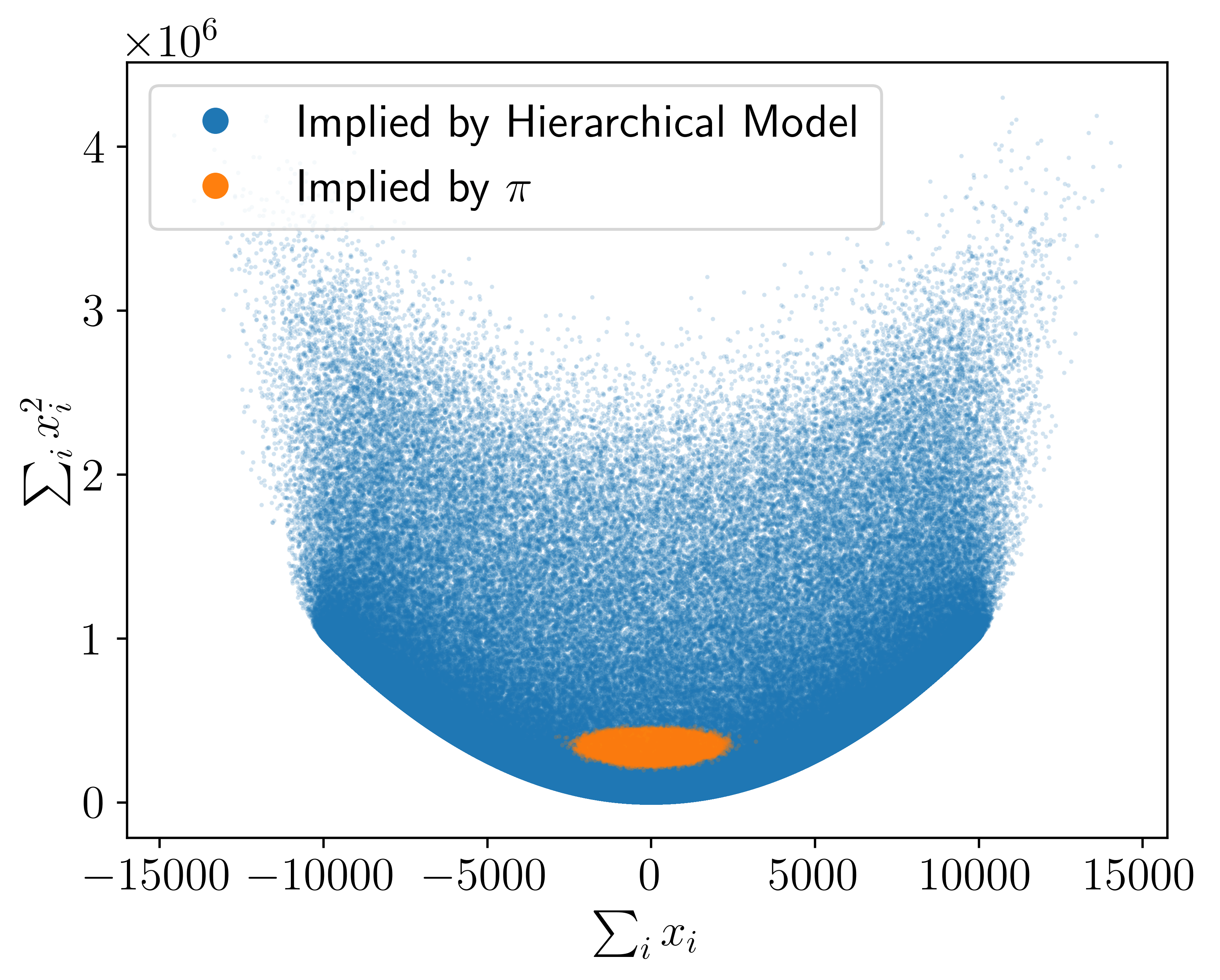}
\caption{The orange distribution is the implied prior for the sum
and sum of squares of 100 quantities with independent Uniform$(-100, 100)$
priors.
The blue distribution is the implied prior using a hierarchical
model, expressing
more appropriate prior uncertainty (approximately uniform in the horizontal
direction and log-uniform vertically) about the sum and the sum of squares.
The U-shape arises from the prior bounds for the hyperparameters.
For completeness, these priors were $\mu \sim \textnormal{Uniform}(-100, 100)$
and $\ln\sigma \sim \textnormal{Uniform}(-5, 5)$ independently.
\label{fig:normal_prior}}
\end{figure}

A hierarchical model can be constructed if we act as if $\mu$ and $\sigma$
are unknown, but have a prior $p(\mu, \sigma)$. The resulting
marginal distribution over $\bx$ would then be given by
\begin{align}
p(\bx) &= \int p(\mu, \sigma)p(\bx \given \mu, \sigma) \, d\mu \, d\sigma \\
    &\propto \int p(\mu, \sigma)\prod_{i=1}^n
            \exp\left(-\frac{1}{2\sigma^2}\left(x_i - \mu\right)^2\right) \, d\mu \, d\sigma \\
    &= \int p(\mu, \sigma)
            \exp\left(\lambda_1 \sum_{i=1}^n x_i + \lambda_2 \sum_{i=1}^n x_i^2\right) \, d\mu \, d\sigma.
\end{align}
Since the entire expression on the right depends on the $\bx$ quantities
only through the `sufficient statistics' $T_1$ and $T_2$, this marginal
prior over $\bx$ is of the form of Equation~\ref{eqn:result2}.
Therefore, by going through this procedure, we have obtained a MaxEnt
distribution over $\bx$. Figure~\ref{fig:normal_prior}
shows an example of the implied prior for the sum and sum of squares
of the $x$ quantities from the uniform prior and the hierarchical model.

\section{Conclusion}
The maximum entropy principle is often applied to expected value or
moment constraints, yielding the canonical distribution as the result.
However, this only applies if we somehow obtain precise values for the
fixed expectations. If we do not have such precise values, a common procedure
is to let the canonical distribution be the {\em conditional}
prior which depends on the unknown values of the Lagrange multipliers
(or their equivalent hyperparameters). A prior for these hyperparameters completes
the model specification.
This is the foundation of `maximum entropy on the mean'
\citep{ali} approach to inverse problems and `superstatistics'
in statistical mechanics \citep{superstatistics}.

This procedure leads to a particular marginal prior for the original
quantities $\bx$ which is {\em not} a canonical distribution, but is a mixture
of canonical distributions. It might seem
that the maximum entropy interpretation
has been lost, but this is not the case. In this contribution, I have demonstrated that the resulting marginal prior for $\bx$ is a MaxEnt
distribution where the implicit constraint is on the marginal prior
for the derived quantities whose expected values were originally fixed.

\section*{Acknowledgements}
I would like to thank Ariel Caticha and Ali Mohammad-Djafari for their
helpful comments on this work.

\end{document}